\documentclass{article}

\usepackage{arxiv}

\usepackage[T1]{fontenc}    
\usepackage{hyperref}       
\usepackage{url}            
\usepackage{booktabs}       
\usepackage{amsfonts}       
\usepackage{nicefrac}       
\usepackage{microtype}      
\usepackage{lipsum}		
\usepackage{graphicx}
\usepackage{natbib}
\usepackage{doi}

\title{A 2020 taxonomy of algorithms inspired on living beings behavior}

\author{ \href{https://orcid.org/0000-0002-8072-6033}{\includegraphics[scale=0.06]{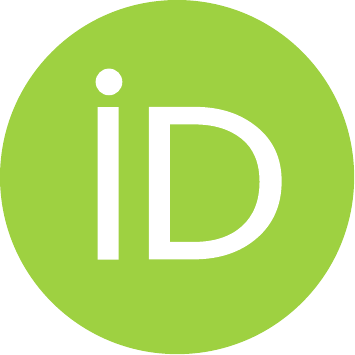}\hspace{1mm}Luis M.~Torres-Trevi\~no}\thanks{https://www.uanl.mx/investigadores/luis-martin-torres-trevino/.} \\
	Posgrado en Ingenier\'ia El\'ectrica\\
	Universidad Aut\'onoma de Nuevo le\'on\\
	Ciudad Universitaria S/N, San Nicol\'as de los Garza, Nuevo Le\'on, M\'exico\\
	\texttt{luis.torrestrv@uanl.edu.mx} \\
}

\renewcommand{\shorttitle}{A 2020 Taxonomy of living beings behavior algorithms}

\hypersetup{
pdftitle={A 2020 taxonomy of algorithms inspired on living beings behavior},
pdfsubject={CS},
pdfauthor={Luis M.~Torres-Trevi\~no},
pdfkeywords={Behavior of living beings, Bio inspired algorithms, Swarm intelligence},
}

\begin{document}
\maketitle

\begin{abstract}
	Taking the role of a computer naturalist, a journey is taken through bio inspired algorithms taking account on algorithms which are inspired on living being behaviors. A compilation of algorithms is made considering several reviews or surveys of bio-inspired heuristics and swarm intelligence until 2020 year. A classification is made considering kingdoms as used by biologists generating several branches for animalia, bacteria, plants, fungi and Protista to develop a taxonomy.
\end{abstract}


\section{Introduction}

Since the emerge of ideas about simulation of life in last decades, several algorithms have been proposed to solve complex problems inspired on nature phenomena; i.e. evolutionary computation or artificial life. A role of a naturalist or biologist is taken with the purpose for studying all living forms in a new ecosystem and trying to make a classification of all discoveries to form a taxonomy of living beings. This role is taken as a computer naturalist to make a compilation of algorithms inspired on behavior of living beings.

There are several bio-inspired algorithms; however, this work focus on actions of living beings like the growth of plants, reproduction of mushrooms, living of bacteria, the individuals behavior of animals, etc.; however, highlights the interactions between individuals of a group of different animals like school of fishes, flock of birds, herd of mammals, or swarm of insects. 

Focusing on algorithms inspired in actions of living beings that belongs to any kingdom of the nature; nevertheless, it is important to locate all algorithms as possible. Only basic algorithms are considered, but derivations, variants and hybrids are omitted; at least,  algorithms which involves an inspiration of any living being. Location of bio-inspired algorithms related with a specific species is made by a review of several papers of surveys which involve nature bio-inspired, swarm intelligence, and metaheuristics algorithms; however, several of these surveys consider different points of view.

It was consider only survey papers from ten years old ago because it is expected a more complete reviews since then. Surveys span in many cases all kind of algorithms; however many of them have been proposed recently; it maybe because the year 2020 is iconic. The most numerous surveys are related with nature inspired algorithms \cite{agarwalsurvey, Siddique2015NatureIC, Sindhuja2018ABS, Krishnaveni2019ASO, Cuevas2020AnIT, YANG2020101104, Sureka2020NatureIM, Odili2018ACR}, or bio-inspired algorithms \cite{Binitha2012ASO, refsurveypazhaniraja2017, Amry2018SurveyOR, DELSER2019220}; specific swarm intelligence algorithms \cite{refparpinelli}  and  metaheuristics or population based algorithms \cite{vanoyeref, refsurveyXing2014, IJSW29497, ISI:000456761600026} Also, a review of applications of swarm intelligence has been made to establish more algorithms. Application of nature inspired algorithms can be found in several areas, not only optimization; however from all natural inspired algorithms, swarm intelligent are the most highlighted.  Dynamic optimization\cite{MAVROVOUNIOTIS20171}, Stock price prediction \cite{ ISI:000543632300001}, feature selection in big data \cite{ISI:000528484400007}, Internet of things \cite{ISI:000525271500182}, Deep learning \cite{ISI:000519571700008}, Automatic clustering \cite{ISI:000517964300134}, VSLI routing problems \cite{ISI:000524665700001}, and image enhancement \cite{ISI:000490604400011}. Comparisons between different algorithms have been made in different works and focus on tests using state of the art benchmarks composed by test multidimensional functions and real problems \cite{refcomparisonBujok, refcomparisonMa, PIOTROWSKI201734,10.1007/s00500-013-1104-9}.


A review of different works about taxonomies was made considering papers which title has the word of {\it taxonomy}. Previous surveys includes taxonomies; however, mostly of them are a brief structures about algorithms, applications or inspirations. A complete set of taxonomies which shows different criteria of classification is made in \cite{SurveyTaxTzanetos2019} where it is presented natural inspired intelligent algorithms which focus on real world problems and applications like traveling salesman problems, operation research problems, energy problems, and other multiple applications. Taxonomy about memory usage, and a chronology  in swarm intelligent metaheuristic algorithms was presented by Shaymah Akram Yasear and Ku Ruhana Ku-Mahamud \cite{reftaxakram2019}.  A complete set of taxonomies of natural and bio-inspired algorithms was presented by Molina et al. \cite{reftaxMolina2020} proposing several classifications like breeding behavior, flying, aquatics, terrestrial, and microorganisms animals. Also includes physics and chemistry based algorithms, social human behavior algorithms, plant based and a miscellaneous source of inspiration algorithms. Saman et al. \cite{JACST29402} present a taxonomy of bio-inspired algorithms focus on optimization where three clases are presented: (i) evolutionary algorithms, (ii) swarm intelligent algorithms, (iii) ecological inspired algorithms. 

This proposal is made to better understanding of algorithms inspired on behavior of living beings; by this way, a taxonomy is proposed based on kingdoms and some statistics are made to establish a global tendency of this promising area.

A complete  taxonomy by behaviors is included where new solutions are generated by differential vector movement and solution creation and finally a taxonomy is made considering more influential algorithms highlighting genetic algorithms, particle swarm optimization, differential evolution, ant colony optimization and artificial bee colony.

\section{Taxonomy explanation}

Specialist are established six kingdoms; however, a traditional Linnaean system of classification of living forms consider five kingdoms: (i) Monera (Bacterias, spirochetes and archaebacterias), (ii) Protista (Algae, protozoans), (iii) Fungi (mushrooms), (iv) Plantae (plants), (v) Animalia (birds, mammals, fishes, insects, etc.). Behavior of animals has been considered in spite of a specific specie is not defined. Virus is not considered a living form; however, a class was established for them.

\subsection{Animalia}
Animalia is the biggest kingdom, made up mammals, fishes, birds, insects, and other animals (other animals are grouped because there are a reduced number of algorithms which represent them). Other algorithms inspired in behavior of animals are included, but these algorithms do no belong to any kingdom. That is the case of virus and algorithms inspired in a general behavior. 

\subsubsection{Birds}

First class of animalia belongs to birds which focus on three forms of behaviors (a) Flocking or migrating, (b) Foraging or Hunting (c) Mating or breeding. Flocking behavior is a general behavior found in birds  ( \cite{FbLref}) and is found in several of them like emperador penguins increase heat radiation  ( \cite{EPCref}), migration birds  ( \cite{DUMAN201265}) or with a \emph{V} configuration used in gooses  ( \cite{WANG20081437}), swallows  ( \cite{SSOAref}) and in pigeons for homing behavior   ( \cite{HPIOref}, ( \cite{PIOref}). A more general form of flocking is made by PSO  ( \cite{psobasic}), and Bird Swarm  ( \cite{doi:10.1080/0952813X.2015.1042530}).  

Algorithms inspired in Foraging or searching for food  are more numerous. Foraging of Hoopoe  ( \cite{ElDosuky2012NewHH}), sparrow searching for worms, seeds  and avoiding predators  ( \cite{doi:10.1080/21642583.2019.1708830}), chickens during food searching  ( \cite{CkSOref}), crows ( \cite{ASKARZADEH20161}) and ravens searching food and sharing information about its location  ( \cite{RROref}) and egyptian vulture  ( \cite{EVtO}) for searching food which is a similar behavior shown by american condor with its  movement patterns ( \cite{ACAref}).

Several algorithms are inspired in hunting which usually is found in predator birds where eagles and Hawks are the favorites to be represented. A flock of hawks cooperatively pounce a pray from different directions  ( \cite{HEIDARI2019849}, \cite{VASCONCELOSSEGUNDO2019119}) Strategies of hunting made by eagles  ( \cite{BESref}, ( \cite{EgSref}) exception is made by eagles perching in higher places  (mountains, cliffs, etc.) for better view for hunting  ( \cite{DBLP:journals/corr/abs-1807-02754}) and collaborative hunting of penguins  ( \cite{PeSOAref}).

Mating and breeding is found in cuckoo bird which is considered a parasite where it found more promises nest to lay eggs ( \cite{5393690}) or mature cuckoos and eggs which try to survives in the population  ( \cite{RAJABIOUN20115508}). Other algorithm use bird mating strategies to increase opportunity of breeding ( \cite{ASKARZADEH20141213}); per example,  laying chicken hatching eggs ( \cite{LCAref}) or during breeding, See-See partridge chicks follow their mother to reach a safe place ( \cite{SSPCOref}) and attraction of females by the satin birds males  ( \cite{SAMAREHMOOSAVI20171}). All algorithms based on birds behavior are shown in Table \ref{tableBirdsBehavior}

\begin{table}[!t]
\centering
\caption{Algorithms based on birds behavior.}
\label{tableBirdsBehavior}
\begin{tabular}{p{0.4\columnwidth}p{0.1\columnwidth}p{0.45\columnwidth}}
\hline
Swarm Algorithms	&	Acronym	&	Authors		\\
\hline						
Andean Condor Algorithm	&	ACA	&		\cite{ACAref}\\
Bald Eagle Search	&	BES	&	\cite{BESref}	\\
Bird Mating Optimizer 	&	BMO	& 	\cite{ASKARZADEH20141213}	\\
Bird Swarm 	&	BrS	&\cite{doi:10.1080/0952813X.2015.1042530}	\\
Chicken swarm optimization	&	CkS	&	\cite{CkSOref}	\\
Crow Search Algorithm	&	CSA	&		\cite{ASKARZADEH20161}	\\
Cuckoo Optimization Algorithm	&	COA	&		\cite{RAJABIOUN20115508}	\\
Cuckoo Search	&	CS	&	\cite{5393690}	\\
Dove Swarm Optimization	&	DSO	&		\cite{SU20092764}	\\
Eagle Perching Optimizer	&	EPO	&	\cite{DBLP:journals/corr/abs-1807-02754}	\\
Eagle Strategies	&	EgS	&		\cite{EgSref}	\\
Egyptian Vulture Optimization 	&	EVO	&	\cite{EVtO}	\\
Emperor Penguins Colony 	&	EPC	&	\cite{EPCref}	\\
Falcon Optimization Algorithm	&	FlOA	&	 \cite{VASCONCELOSSEGUNDO2019119}	\\
Flock by Leader	&	FbL	&	\cite{FbLref}	\\
Goose Team Optimization	&	GTO	&	 \cite{WANG20081437}	\\
Harris'  Hawk Optimization	&	HrHO	&. \cite{HEIDARI2019849}	\\
Heterogeneous Pigeon-inspired Optimization 	&	HPIO	&	 \cite{HPIOref}	\\
Hoopoe Heuristic Optimization 	&	HHO	&		\cite{ElDosuky2012NewHH}	\\
Laying Chicken Algorithm	&	LCA	&		\cite{LCAref}	\\
Migrating Birds Optimization 	&	MBO	&	\cite{DUMAN201265}	\\
Owl Search Algorithm	&	OSA	&	 \cite{OSAref}	\\
Particle Swarm Optimization	&	PSO	&	 \cite{psobasic}	\\
Penguins Search Optimization Algorithm 	&	PnSO	&	\cite{PeSOAref}	\\
Pigeon Inspired Optimization	&	PIO	&		\cite{PIOref}	\\
Raven Roosting Optimization Algorithm 	&	RRO	& \cite{RROref}	\\
Satin Bowerbird Optimizer	&	SBO	&		\cite{SAMAREHMOOSAVI20171}	\\
SeaGull Optimization Algorithm	&	SgOA	&	 	\cite{DHIMAN2019169}	\\
See-See Partridge Chicks Optimization	&	SPCO	&	\cite{SSPCOref}	\\
Sooty Tern Optimization Algorithm	&	STOA	&	\cite{DHIMAN2019148}	\\
Sparrow Search Algorithm & SpSA &	 	\cite{doi:10.1080/21642583.2019.1708830}	\\
Swallow Swarm Optimization Algorithm & SSOA	&  \cite{SSOAref}	\\
\hline
\end{tabular}
\end{table}

\subsubsection{Fishes}
Second class of animalia kingdom belongs to fishes. Almost all behaviors are for hunting and foraging. Catfish effect is used when a school of sardines in a water tank are motivated to move when a catfish (a natural predator) is included in the tank  ( \cite{CHUANG201112699}). Shark uses an improved version of a fish search algorithm ( \cite{refFSA}) to increase exploration procedure for hunting  ( \cite{HERSOVICI1998317}). Other ability of sharks for hunting is to use smell sense  ( \cite{doi:10.1002/cplx.21634}). Yellow Saddle Goatfishes forms schools for hunting taking two roles, chase to pursuit preys and blockers to avoid escapes ( \cite{ZALDIVAR20181}).

 A similar behavior is taken by sailfish for hunting sardines  ( \cite{SHADRAVAN201920}). Mouth brooders fishes take care of their offsprings by using their mouth as a form of protection again environmental conditions and predators ( \cite{JAHANI2018987}). The Great Salmon Run is a migration for mating where salmon return from the sea to the mountains where they born overcoming all kind of menaces ( \cite{RefGSR}). All algorithms based on fishes behavior are shown in Table \ref{tableFishesBehavior}

\begin{table}[!t]
\centering
\caption{Algorithms based on fishes behavior.}
\label{tableFishesBehavior}
\begin{tabular}{p{0.4\columnwidth}p{0.1\columnwidth}p{0.45\columnwidth}}
\hline
Swarm Algorithms	&	Acronym	&	Authors		\\
\hline						
Catfish Optimization Algorithm	&	CtOA	&	\cite{CHUANG201112699}	\\
Circular Structure of Puffer Fish	&	CSPF	& \cite{8357123}	\\
Fish School Search	&	FSS	&		\cite{FSSref}	\\
Fish Swarm Algorithm 	&	FSA	&		\cite{refFSA}	\\
Great Salmon Run 	&	GSR	&	\cite{RefGSR}	\\
Mouth Breeding Fish Algorithm	&	MBFA	&		\cite{JAHANI2018987}	\\
Shark Smell Optimization 	&	ShSO	&		\cite{doi:10.1002/cplx.21634}	\\
Shark-Search Algorithm	&	SSA	&	 \cite{HERSOVICI1998317} \\
The Sailfish Optimize 	& SFO	&  \cite{SHADRAVAN201920} \\
Yellow Saddle Goat Fish & YSGF &  \cite{ZALDIVAR20181} \\
\hline
\end{tabular}
\end{table}

\subsubsection{Insects}

Third class belongs to insects where several algorithms imitate behavior of social insects; however, there are a redundancy because  all insects presents the same behaviors as other animals: (i) Foraging, (ii) hunting, (iii) mating but rarely breading, and (iv) other specific behaviors. Ants, bees and wasp presents complex social behaviors which they are very attractive to develop algorithms for emulation. Behavior of ants was first imitated to simulate the use of pheromones to establish potential sources of food or to begin  a foraging process ( \cite{Dorigo,DorigoThesis}), with a random walk procedure to improve exploration  ( \cite{10.1007/11758501_117}, ( \cite{AJNU}). Other algorithms inspired in ants focus on behavior based on hunting mechanisms ( \cite{MIRJALILI201580}). Termites have exhibit the same collective behavior  ( \cite{5507009}) and cooperative behavior of hill building  ( \cite{ZUNGERU20121901}). Bees are social insect with complex mechanism of communications and have a sort of several behaviors for foraging, mating, avoiding predators, etc. This behavior is another very attractive task to be inspired to develop swarm algorithms. Colony bees foraging made artificial is the most popular  ( \cite{BeesAref}, \cite{10.1145/1569901.1569906},  \cite{BLAref}, \cite{BCOref}, \cite{HSFref}, \cite{doi:10.1177/105971230401200308},\cite{aeb9f70a9780471e8d85032b34b29f5}, \cite{6855842}) or mating process in bees where queen flights high from the nest followed by the males  (\cite{934391}) or evolution of bees  (\cite{1192242}, \cite{VBAref})

Behavior of individual insects with social interaction considers several tasks like navigation, foraging or escape from predators. This behavior have been inspired from dragonflies ( \cite{DAref}), approaching or hovering over food sources of fruitful flies  ( \cite{PAN201269}, \cite{6933060}), worms foraging  ( \cite{Arnaout2014WormOA}), foraging and mating of butterflies ( \cite{7495523}, \cite{RBOAref, ABtOref}) or only focus on males butterflies to mate-location behavior  ( \cite{BtOAref}). Long-horn beetle for searching food and avoid predators  ( \cite{DBLP:journals/corr/abs-1710-10724}), bark beetles searching for food and a nest ( \cite{PBAref}), and foraging of seven spot ladybirds  ( \cite{SLOref}). Attraction and flashing behaviors of fireflies ( \cite{FFAref}) or lampyridaeis to produce light by bioluminescense for attraction of preys or mating  ( \cite{BSOref}). 
Foraging and social interaction of spiders to establish the positions of preys based on detection of vibrations of the spider web  ( \cite{CUEVAS20136374}, \cite{YU2015614}) or black widow spiders bizarre mating  ( \cite{HAYYOLALAM2020103249}). Wasps interact and allocate the tasks required in nest  ( \cite{WSOref},\cite{6294123}). Only a collective behavior of aggregation has been represented in several algorithms, inspired on glowworms  ( \cite{GSOref}), grasshopper and locust swarm  ( \cite{SAREMI201730}, \cite{4983152}), cockroaches  ( \cite{4668317}, \cite{CSOref}). Migration made by monarch buttlerflies  ( \cite{MBOref}) Attraction of females crickets by  sound emitted by males chirping of wings  ( \cite{CBBECref}) or mating in bumblebees  ( \cite{BBMO}) and mayflies ( \cite{ZERVOUDAKIS2020106559}). Different behaviors of Mosquitos have been used considering seeking a host ( \cite{GMHSAref}), selection of habitat lo lay eggs ( \cite{MINHAS20114614}), and to find a hole in a net ( \cite{7754783}). All algorithms based on insects behavior are shown in Tables \ref{tableInsectBehavior1} and  \ref{tableInsectBehavior2}.

\begin{table}[!t]
\centering
\caption{Algorithms based on insects behavior (part 1).}
\label{tableInsectBehavior1}
\begin{tabular}{p{0.4\columnwidth}p{0.1\columnwidth}p{0.45\columnwidth}}
\hline
Swarm Algorithms	&	Acronym	&	Authors		\\
\hline						
Ant Colony Optimization	&	ACO	&		\cite{Dorigo,DorigoThesis}	\\
Ant Lion 	&	ALO	&	S. Mirjalili 	\cite{MIRJALILI201580}	\\
Artificial Bee Colony	&	ABC	&		\cite{Karaboga2007}	\\
Artificial Beehive Algorithm	&	ABA	&		\cite{ABAref}	\\
Artificial Butterfly Optimization	&	ABO	&	\cite{ABtOref}	\\
Bee Colony Optimization	&	BCO	&		\cite{BCOref}	\\
Bee Colony-Inspired Algorithm	&	BciA	&	\cite{10.1145/1569901.1569906}	\\
Bee Swarm Algorithm	&	BeeSA	& \cite{doi:10.1080/0952813X.2015.1042530}	 		\\
Bee Swarm Optimization	&	BSO	&	 \cite{BSOref}	\\
Bee System	&	BS	&		\cite{aeb9f70a9780471e8d85032b34b29f5}	\\
Bees Algorithm	&	BA	&		\cite{BeesAref}	\\
Bees Life Algorithm	&	BLA	&		\cite{BLAref}	\\
Bettle Antennae Search Algorithm	&	BASA	&	\cite{DBLP:journals/corr/abs-1710-10724}	\\
Bioluminescent Swarm Optimization	&	BSO	&		\cite{BSOref}	\\
Black Widow Optimization	&	BWO	&		\cite{HAYYOLALAM2020103249}	\\
Bumble Bees Mating Optimization & BBMO &  \cite{BBMO} \\
Butterfly Optimization Algorithm	&	BtOA	&	\cite{BtOAref}	\\
Buttlerfly Optimizer	&	BTO	& \cite{7495523}	\\
Coachroach Swarm Optimization	&	CrSO	&		\cite{CSOref}	\\
Cricket Behaviour-Based Evolutionary	&	CBBE	&		\cite{CBBECref}	\\
Dispersive Flies Optimization	&	DFO	&		\cite{6933060}	\\
Dragonfly Algorithm	&	DA	&		\cite{DAref}	\\
Firefly Algorithm	&	FA	&		\cite{FFAref}	\\
Fruitfly Algorithm	&	FtA	&		\cite{PAN201269}	\\
\hline
\end{tabular}
\end{table}

\begin{table}[!t]
\centering
\caption{Algorithms based on insects behavior(part 2).}
\label{tableInsectBehavior2}
\begin{tabular}{p{0.4\columnwidth}p{0.1\columnwidth}p{0.45\columnwidth}}
\hline
Swarm Algorithms	&	Acronym	&	Authors		\\
\hline						
Glowworm Swarm Optimization Algorithm	&	GSO	&		\cite{GSOref}	\\
Grasshopper Optimisation Algorithm 	&	GOA	&	 	\cite{SAREMI201730}	\\
Group Mosquito host-seeking algorithm	&	GMHA	&	 \cite{GMHSAref}	\\
Honey Bee Behavior	&	HBB	&	\cite{HBBref}	\\
Honey Bees Optimization	&	HBO	&		\cite{doi:10.1177/105971230401200308}	\\
Honey-bees Mating Optimization Algorithm	&	HBMOA	&	\cite{934391}	\\
Honeybee Social Foraging	&	HSF	&		\cite{HSFref}	\\
Locust Swarm	&	LS	&	Stephen Chen 	\cite{4983152}	\\
Mayfly Optimization Algorithm &   MayfOA  &  \cite{ZERVOUDAKIS2020106559}  \\
Monarch Butterfly Optimization	&	MnBO	&		\cite{MBOref}	\\
Mosquito Fly Optimization	&	MqFO	&		\cite{7754783}	\\
Mosquitos Oviposition 	&	MOX	&	\cite{GMHSAref}	\\
Moth-flame Optimization Algorithm 	&	MfOA	&	\cite{MIRJALILI2015228}	\\
Mox Optimization Algorithm	&	MOA	&	\cite{MINHAS20114614}	\\
OptBees	&	OB	&		\cite{6855842}	\\
Pity Beetle Algorithm	&	PBA	&	\cite{PBAref}	\\
Queen-bees Evolution 	&	QBE	&	\cite{1192242}	\\
Regular Butterfly Optimization Algorithm	&	RBOA	&	\cite{RBOAref}	\\
Roach Infestation Optimization 	&	RIO	&	 \cite{4668317}	\\
Seven-spot Ladybird Optimization 	&	SLO	&		\cite{SLOref}	\\
Social Spider Algorithm	&	ScSA	&		\cite{CUEVAS20136374}	\\
Social Spider Optimization	&	SSO	&	\cite{YU2015614}	\\
Termite Colony Optimization 	&	TCO	&	  \cite{5507009}	\\
Termite Hill Algorithm	&	THA	&	  \cite{ZUNGERU20121901}	\\
U-Turning Ant Colony Optimiza-tion 	&	U-TACO	&	  	\cite{AJNU}	\\
Virtual Ants Algorithm	&	VAA	&	 \cite{10.1007/11758501_117}	\\
Virtual Bees Algorithm	&	VBA	&	\cite{VBAref}	\\
Wasp Colonies Algorithm	&	WCA	&	 	\cite{6294123}	\\
Wasp Swarm Optimization 	&	WSO	&	 	\cite{WSOref}	\\
Worm Optimization 	&	WO	&	 \cite{Arnaout2014WormOA}	\\
\hline
\end{tabular}
\end{table}

\subsubsection{Mammals}

Algorithms inspired in behavior of mammals are the second more numerous members of animalia kingdom since the first one are insects. A classification of behaviors is proposed again dividing algorithms for foraging, hunting and specific behaviors. Hunting behavior of some mammal predators has been represented in several algorithms, per example colony of bats using echolocation to find a prey  ( \cite{batalgref}, \cite{BIref}, \cite{DBLP:journals/corr/abs-1211-0730}), canides as dogs ( \cite{AWDAref}, \cite{vanoyeref}, \cite{WDPref}), hyenas  ( \cite{DHIMAN201748}), coyotes  ( \cite{PIEREZAN2019111932}, \cite{8477769}) and the wolfs  ( \cite{MIRJALILI201446}, \cite{WCOref}, \cite{4438476}, \cite{6360147})  and felines like lion ( \cite{10.1016/j.jcde.2015.06.003}) jaguar ( \cite{JAref}), cheeta  ( \cite{IJET14616}) and cats  ( \cite{10.1007/978-3-540-36668-3_94})

Sea lions hunting behavior using whiskers to detect the prey ( \cite{LPOref}). Complex hunting and exploration mechanism made by polar bears ( \cite{PBOAref}). Rummaging food and remembering food source of racoons ( \cite{8552661}) 

Aggregation of mammals for foraging, searching or feeding behavior  is very common  in rumiantes like buffalo  ( \cite{ABOref}), bisons ( \cite{BBAref}), camel  ( \cite{CAref}), and the elephants ( \cite{7383528}, \cite{7381893}), the squirrels  ( \cite{JAIN2019148}), flying squirrel  ( \cite{FSOref}) and searching behavior of the donkeys to find routes  ( \cite{SHAMSALDIN2019562}).

An artificial selection of the best individual, a form of extreme elitism is made in bull to be selected to generate strong and better offsprings ( \cite{BullOAref}. A similar mechanism of competition for mating is made by Reed Deer ( \cite{RDAref}) or lions prides behavior ( \cite{LPOref}) where strongest males have high probabilities for mating and eliminate other competitors inclusive theirs offsprings.

Marine mammals have serve as inspiration to several algorithms specially dolphins and whales. The use o echolocation ( \cite{KAVEH201353}) or hunting strategies where information is shared  between dolphins is commonly used as inspiration ( \cite{Zhao_2015jaciii}, \cite{5209016},\cite{DpdOref}) The same behavior is used in gray whales for foraging strategies using burbles ( \cite{MIRJALILI201651}) or searching and hunting like sperm whale ( \cite{EBRAHIMI2016211}) or killer whale ( \cite{KWAref}) 
 
A more complex behavior can be observed in several primates like spider monkey  ( \cite{SMOref}), blue monkeys  ( \cite{BMAref}), ageist monkeys which age could bring a difference for better performance to solve problems  ( \cite{SHARMA201658}) and sexual behavior with individual intelligence of chimps ( \cite{KHISHE2020113338}). Humans are considered primates and have the same behavior but in a more sophisticated way to form tribes  ( \cite{5658772}), groups ( \cite{THAMMANO20101628}, \cite{5357838}) or aggregates as crowds ( \cite{WACref}) (More complex behaviors unique for humans are not considered here because it is very difficult to be expressed by other species) All algorithms based on mammals behavior are shown in Tables \ref{tableMammalBehavior1} and \ref{tableMammalBehavior2}.

\begin{table}[!t]
\centering
\caption{Algorithms based on mammals behavior (part 1).}
\label{tableMammalBehavior1}
\begin{tabular}{p{0.4\columnwidth}p{0.1\columnwidth}p{0.45\columnwidth}}
\hline
Swarm Algorithms	&	Acronym	&	Authors		\\
\hline						
African Buffalo Optimization 	&	ABO	&		\cite{ABOref}	\\
African Wild Dog Algorithm 	&	AWDA	&	\cite{AWDAref}	\\
Ageist Spider Monkey Optimization	&	ASMO	&  \cite{SHARMA201658}	\\
Artificial Tribe Algorithm	&	ATA	&	\cite{5658772}	\\
Bat Algorithm	&	BA	&	 	\cite{batalgref}	\\
Bat Intelligence	&	BI	&		\cite{BIref}	\\
Bat Sonar Algorithm	&	BSA	&	\cite{DBLP:journals/corr/abs-1211-0730}	\\
Bison Behavior Algorithm	&	BBA	& \cite{BBAref}	\\
Blind, Naked Mole-rats Algorithm 	&	BNMR	&		\cite{TAHERDANGKOO20131}	\\
Bull Optimization Algorithm 	&	BllOA	&  \cite{BullOAref}	\\
Camel Algorithm 	&	CA	&	 \cite{CAref}	\\
Cat Swarm Optimization	&	CSO	&	 \cite{10.1007/978-3-540-36668-3_94}	\\
Cheetah Chase Algorithm	&	CCA	&	\cite{IJET14616}	\\
Chimp Optimization Algorithm & COA &  \cite{KHISHE2020113338} \\
Coyote Optimization Algorithm	&	CyOA	&	\cite{8477769}	\\
Cultural Coyote Optimization Algorithm	&	CCOA	&		\cite{PIEREZAN2019111932}	\\
Dolphin Echolocation 	&	DphE	&		\cite{KAVEH201353}	\\
Dolphin Herd Algorithm	&	DHA	&	\cite{Zhao_2015jaciii}	\\
Dolphin Partner Optimization 	&	DPO	&		\cite{5209016}	\\
Dolphin Pod Optimization	&	DpdO	&		\cite{DpdOref}	\\
Donkey and Smuggler Optimization Algorithm	&	DsOA	&	 \cite{SHAMSALDIN2019562}	\\
Elephant Herding Optimization 	&	EHO	&		\cite{7383528}	\\
Elephant Search Algorithm 	&	ESA	&	 \cite{7381893}	\\
Feral, Dogs Herd algorithm	&	FDHA	&		\cite{vanoyeref}	\\
Flying Squirrel Optimizer	&	FSO	&	 \cite{FSOref}	\\
Gray Wolf Optimizer	&	GWO	&	\cite{MIRJALILI201446}	\\
Human Group Formation	&	HGF	&	\cite{THAMMANO20101628}	\\
Human-inspired Algorithm 	&	HIA	&	 \cite{5357838}	\\
Jaguar Algorithm with Learning Behavior 	&	JA	&	\cite{JAref}	\\
\hline
\end{tabular}
\end{table}

\begin{table}[!t]
\centering
\caption{Algorithms based on mammals behavior (part 2).}
\label{tableMammalBehavior2}
\begin{tabular}{p{0.4\columnwidth}p{0.1\columnwidth}p{0.45\columnwidth}}
\hline
Swarm Algorithms	&	Acronym	&	Authors		\\
\hline						
Killer whale algorithm	&	KWA	&	 \cite{KWAref}	\\
Leader of Dolphin Herd Algorithm	&	LDHA	&	\cite{Zhao_2015jaciii}	\\
Lion Optimization Algorithm 	&	LOA	&		\cite{10.1016/j.jcde.2015.06.003}	\\
Lion Pride Optimizer	&	LPO	&		\cite{LPOref}	\\
Meerkats Clan Algorithms	&	MCA	&		\cite{MCAref}	\\
Meerkats Inspired Algorithm	&	MIA	&		\cite{Klein2018MeerkatsinspiredAF}	\\
Monkey Search 	&	MS	&		\cite{doi:10.1063/1.2817338}	\\
Naked Moled Rat	&	NMR	&	 	\cite{NMRAref}	\\
Polar Bear Optimization Algorithm	&	PBOA	&		\cite{PBOAref}	\\
Raccoon Optimization Algorithm	&	ROA	&  \cite{8552661}	\\
Rats Herd Algorithm	&	RATHA	&	\cite{vanoyeref}	\\
Red Deer Algorithm	&	RDA	&		\cite{RDAref}	\\
Rhino Herd Behavior	&	RHB	&	\cite{RHBref}	\\
Sea Lion Optimization Algorithm	&	SLnOA 	&		\cite{LPOref}	\\
Sheep Flocks Heredity Model 	&	SFHM	&		\cite{816603}	\\
Sperm Whale Algorithm 	&	SWA	&	\cite{EBRAHIMI2016211}	\\
Spider Monkey Optimization 	&	SMO	&		\cite{SMOref}	\\
Spotted Hyena Optimizer	&	SHO	&		\cite{DHIMAN201748}	\\
Squirrel Search Algorithm	&	SqSA	&	\cite{JAIN2019148}	\\
The Blue Monkey 	&	TBM	&	\cite{BMAref}	\\
Whale Optimization Algorithm	&	WOA	&		\cite{MIRJALILI201651}	\\
Wild Dog Packs	&	WDP	&	 \cite{WDPref}	\\
Wildebeests Herd Optimization	&	WHO	&	\cite{doi:10.1142/S0218001419590171}	\\
Wisdom of Artificial Crowds	&	WAC	&	\cite{WACref}	\\
Wolf Colony Algorithm	&	WlCA	&	\cite{WCOref}	\\
Wolf Pack Search	&	WPSA	&	\cite{4438476}	\\
Wolf Search Algorithm 	&	WSA	&	\cite{6360147}	\\
\hline
\end{tabular}
\end{table}

\subsubsection{Others living beings from animalia}

Fifth class is a group of several classes of algorithms which are inspired in behavior of other animals or phylum with a low number of representations.
These groups of animals like mollusks where cuttlefish is a celopod which establish a correct color to mimic on the environment (\cite{CFAref}), krill living for foraging and reproduction (\cite{GANDOMI20124831}), bed formations by  leisurely locomotion of mussels for survival (\cite{MWOref}) or swarm behavior of salps to navigate and forage (\cite{MIRJALILI2017163}). 

Different corals fighting with other coral for space on the reef (\cite{CROAref}) and Zooplankton which integrates several species of animals, larvae, and other organisms (\cite{vanoyeref}). Amphibius have been considered like behavior of japanese tree frogs where for mating, males frogs calls females through desynchronization calls (\cite{5227977}). Others frogs interchange elements of cultural information called memes; some of them are hosts an others are carriers. This behavior is used  to improve exploration and exploitation for foraging (\cite{doi:10.1080/03052150500384759}). Mating by polymorphic of side-blotched lizards (\cite{MACIELC2020106039}) or simply foraging of frogs (\cite{JFOref}).
Finally, algorithms for optimization based on animal behavior are inspired in a general performance of animals without a given specification of the species or it can be applied to any species. This could be assumed to humans only but; in some cases, it could be presented in another animals like ants, bees, wolfs, or whales. Complex behavior like simbiosis (\cite{CHENG201498}), formation of structures (\cite{4341350}) or a high order of behavior like cooperation (\cite{CIVICIOGLU201358}) and competition (\cite{7057107}) is used. However, other algorithms emphasize collective behaviors trying to form a general swarm algorithm like Universal Swarm Optimizer (\cite{USoref}), hierarchy formation (\cite{HSMref}) or to pursuit an objective, like search (\cite{1688455}) aggregation (\cite{5194080}, \cite{CABref}, \cite{CUI2006505}), hunting a prey (\cite{5379451}, (\cite{PrPAref}) or migration (\cite{AMOref}, \cite{PMAref}, \cite{zelinka}).  All algorithms based on a general animal based behavior are shown in Table \ref{tableAnimalBehavior}).

\begin{table}[!t]
\centering
\caption{Algorithms based on animal behavior.}
\label{tableAnimalBehavior}
\begin{tabular}{p{0.4\columnwidth}p{0.1\columnwidth}p{0.45\columnwidth}}
\hline
Swarm Algorithms	&	Acronym	&	Authors		\\
\hline						
Animal Migration Optimization Algorithm 	&	AMO	&  \cite{AMOref}	\\
Artificial Cooperative Search Algorithm	&	ACSA	&		\cite{CIVICIOGLU201358}	\\
Artificial Searching Swarm Algorithm 	&	ASSA	&		\cite{5194080}	\\
Collective Animal Behavior 	&	CAB	&	 \cite{CABref}	\\
Competition Over Resources	&	COR	&	\cite{7057107}	\\
Flocking-based Algorithm	&	FBA	&	 \cite{CUI2006505}	\\
Good Lattice Swarm Optimization	&	GLSO	&	\cite{4341350}	\\
Group Search Optimizer 	&	GpSO	&	 \cite{1688455}	\\
Hierarchical Swarm Model 	&	HSM	&	 \cite{HSMref}	\\
Hunting Search Optimization 	&	HSO	& 	\cite{5379451}	\\
Population Migration Algorithm	&	PMA	&	\cite{PMAref}	\\
Predator Prey Algorithm	&	PdPA	&	\cite{PrPAref}	\\
Self-Organized Migration Algorithm	&	SOM &	\cite{CHENG201498}	\\
Universal Swarm Optimizer	&	USO	&  \cite{USoref}	\\
\hline
\end{tabular}
\end{table}

\subsection{Bacteria}

In this subsection are combined two kingdoms to consider bacteria and archaebacterias. Bacterias have several complex behaviors for survival; all algorithms are inspired a life cycling of bacterias, from foraging, reproduction, elimination (\cite{BGAFref},\cite{BtCOref}, \cite{4983241}, \cite{1004010}, \cite{4631222},\cite{SBOref})  communication by chemotaxis (\cite{985689}). Some bacterias orient and swim through geomagnetic fields or by magnetostatic for foraging (\cite{6615185}). All algorithms based on bacterial behavior are shown in Table \ref{tableBacterialBehavior})

\begin{table}[!t]
\centering
\caption{Algorithms based on bacterial behavior.}
\label{tableBacterialBehavior}
\begin{tabular}{p{0.4\columnwidth}p{0.1\columnwidth}p{0.45\columnwidth}}
\hline
Swarm Algorithms	&	Acronym	&	Authors		\\
\hline						
Bacterial GA Foraging	&	BGAF	&	 \cite{BGAFref}	\\
Bacterial Chemotaxis Algorithm	&	BCA	&	\cite{985689}	\\
Bacterial Colony Optimization	&	BcCO	&	 	\cite{BtCOref}	\\
Bacterial Evolutionary Algorithm 	&	BEA	&		\cite{4983241}	\\
Bacterial Foraging Algorithm	&	BFA	&	K.M. Passino 	\cite{1004010}	\\
Bacterial Swarming	&	BS	&		\cite{4631222}	\\
Magnetotactic Bacteria Optimization Algorithm 	&	MBOA	&		\cite{6615185}	\\
SuperBug Algorithms	&	SBA	&	\cite{SBOref}	\\
\hline
\end{tabular}
\end{table}

\subsection{Fungi}

There is only one algorithm that belong to this kingdom which it is inspired in mushrooms, specifically in their reproduction and growth mechanisms made in nature. Mushrooms expulse spores to discover good living conditions for growing and develop colonies (\cite{8477837}).

\subsection{Plants}

Growing of plants as an adaptive phenomenon is represented in several algorithms like transmission mode of bean seeds (\cite{BeanOAref}). The growing process of trees where branches growth toward betters positions by phototropism and photosynthesis (\cite{APOAref}). In a more sophisticated way, a complete process of a forest growth has been represented where trees can live more time in specific zones and others trees can not (\cite{GHAEMI20146676}). Some Seeds falls near of these trees but others are taken to another promising zones. A similar process where an artificial plant growth is represented including growth of leaves and spatial branching by phototropism  (\cite{4659682}). Strawberry plant which try to find better conditions of water, light and nutrients investing more in promising  near neighbor spots has inspired some algorithms (\cite{SAref}). The colonization in a cropping of invasive weeds  (\cite{MEHRABIAN2006355}). Seeds in more fertile soils, grow more and produce more seeds on paddy fields (\cite{PFAref}). Other mechanisms used by plants to defend against different predators since fungus, to insects or herbivores (\cite{SDTPref}). Pollination process of the flowers represents different strategies followed by the plants to increase its reproduction (\cite{GHAEMI20146676}), and the growth of roots to search high nutrients (\cite{RGOref}, \cite{LABBI2016298}). A representation of the growth of root tree for searching wet soil (\cite{RRAref}). All algorithms based on plants mechanisms are shown in Table \ref{tablePlantBehavior}

\begin{table}[!t]
\centering
\caption{Algorithms based on plants behavior.}
\label{tablePlantBehavior}
\begin{tabular}{p{0.4\columnwidth}p{0.1\columnwidth}p{0.45\columnwidth}}
\hline
Swarm Algorithms	&	Acronym	&	Authors		\\
\hline						
Artificial Plant Optimization Algorithm	&	APOA	&		\cite{APOAref}	\\
Bean Optimization Algorithm	&	BOA	&		\cite{BeanOAref}	\\
Flower pollination Algorithm	&	FPA	&		\cite{Yang_2012}	\\
Forest Optimization Algorithm	&	FOA	&		\cite{GHAEMI20146676}	\\
Invasive Weed Optimization	&	IWO	&		\cite{MEHRABIAN2006355}	\\
Natural Forest Regeneration	&	NFR	&	\cite{NFRref}	\\
Paddy Field algorithm & PFA & \cite{PFAref} \\
Plant Growth Optimization	& PGO & \cite{4659682} \\
Plant Propagation Algorithm	&	PPA	&	\cite{PPAref}	\\
Root Growth Optimizer	&	RGO	&	\cite{RGOref}	\\
Rooted Tree Optimization Algorithm	&	RTOA	&		\cite{LABBI2016298}	\\
Runner-Root Algorithm & RRA & \cite{RRAref} \\
Sapling Growing up Algorithm	&	SGUA	&		\cite{SGUAref}	\\
Seed Based Plant Propagation Algorithm	&	SBPPA	&	\cite{SBPPAref}	\\
Self-defense Techniques of the Plants	&	SDTP	&	\cite{SDTPref}	\\
Strawberry Algorithm	&	SA	&	\cite{SAref}	\\
\hline
\end{tabular}
\end{table}

\subsection{Protista}

Living forms that belong to Protista kingdom has inspired some algorithms, i.e.  Adaptation and movement of micro algaes (\cite{UYMAZ2015153}).  Slime mould movement as propagation wave  through paths for food by bio-oscillatory movement has inspired some algorithms (\cite{4668295}). A similar behavior is made by Amoeboid Organisms (\cite{6243094}).

\subsection{Viruses}

Finally, virus do not belong to any kingdom; however, they represents their own kingdom according with some authors; however they have an iteration with species of other  kingdoms. The capabilities to infect successfully cells and produce new viruses is represented in (\cite{CORTES20082840,LI201665,doi:10.1080/0305215X.2014.994868, JADERYAN2016596}). A more recent algorithm inspired in viruses behavior is Coronavirus optimization algorithm which is based on propagation of COVID-19 illness (\cite{coronavirus})

\subsection{Who is who in living being inspired algorithms}
\begin{figure}
\centering
\includegraphics[width=18cm]{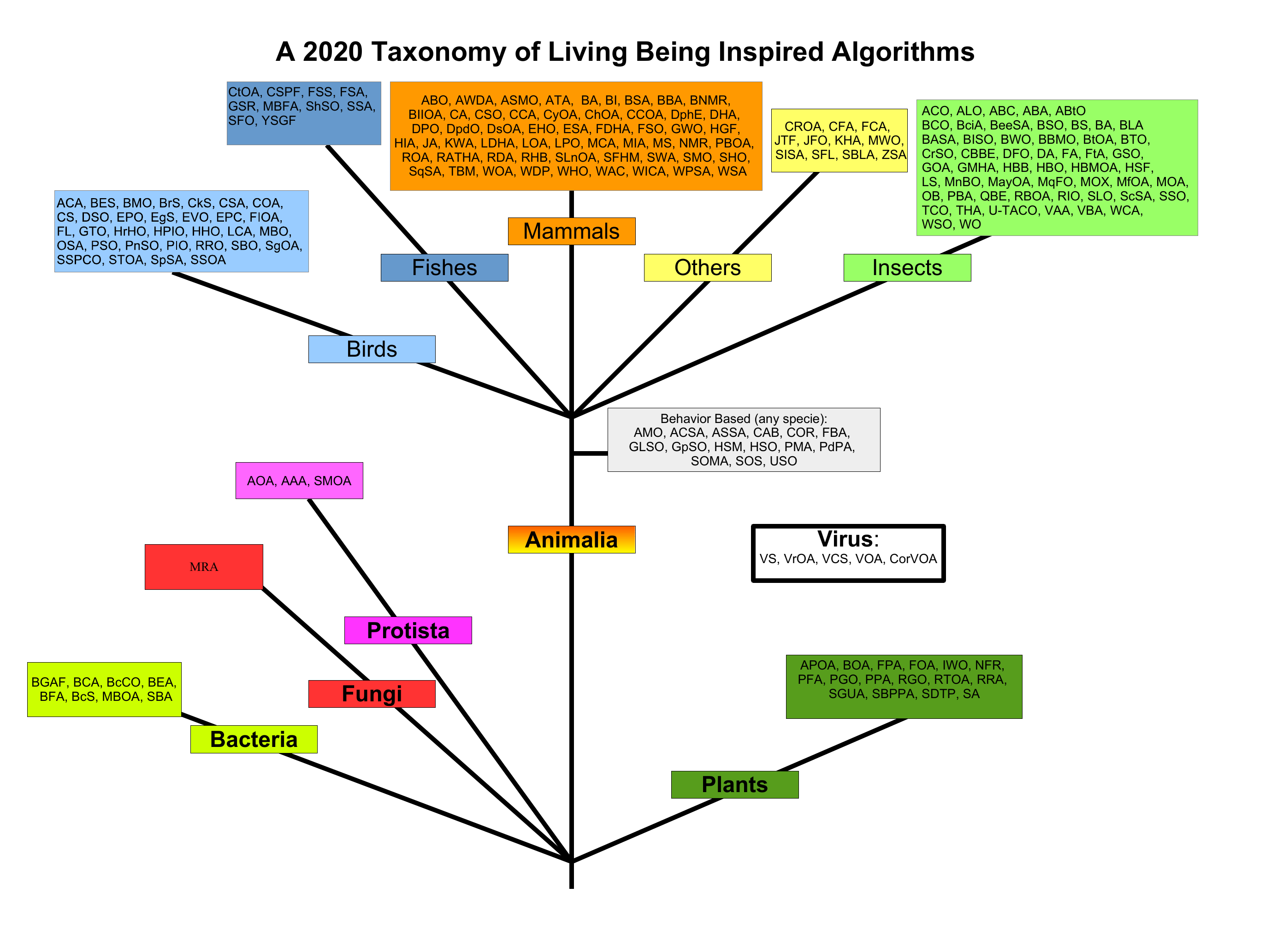}
\caption{Taxonomy of algorithms inspired on behavior of living beings.}
\label{taxonomy}
\end{figure}

\begin{figure}[!t]
    \centering
    \includegraphics[width=15cm]{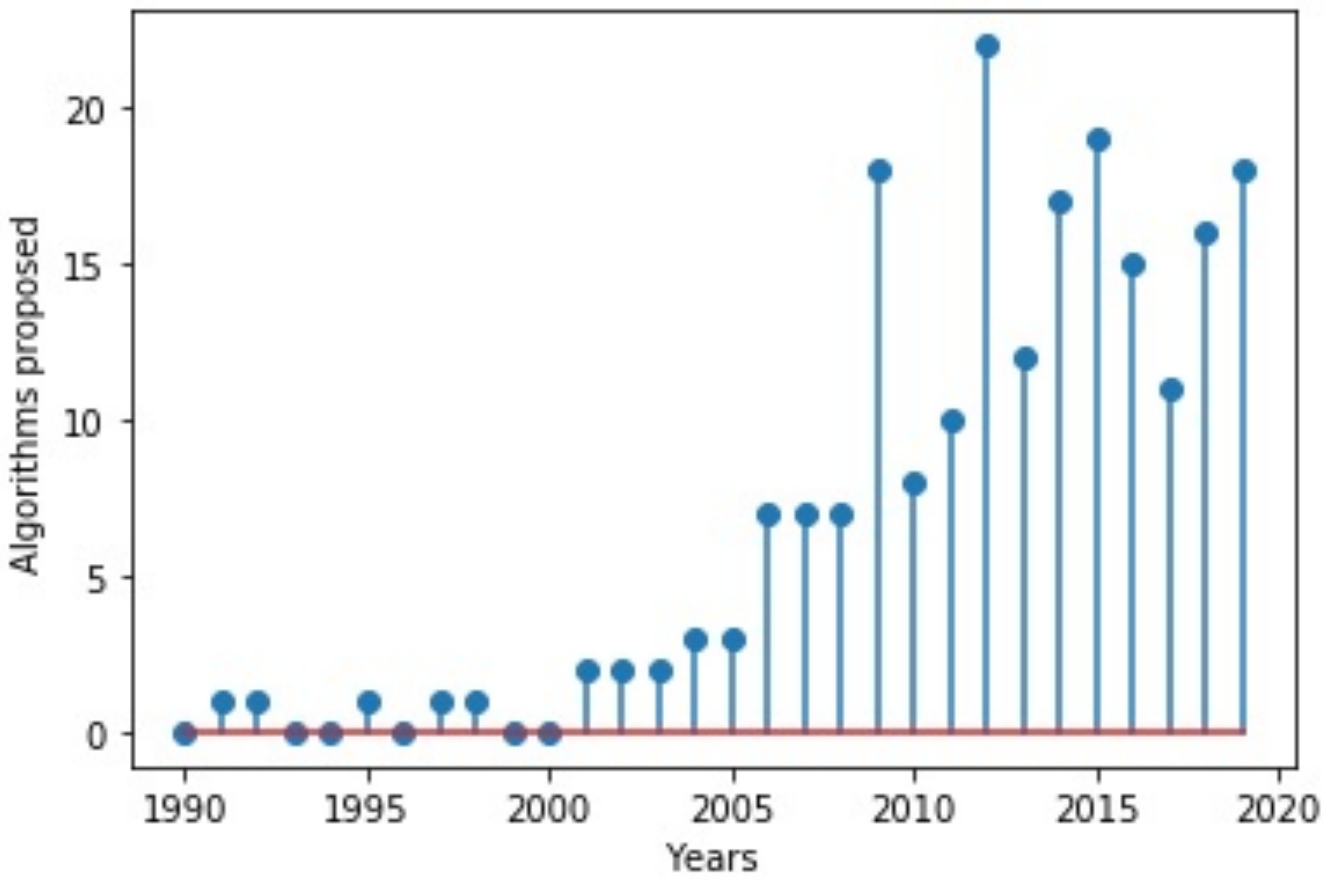}
		\caption{Quantity of new proposals for year.}
		\label{algyear}
\end{figure}

Complete taxonomy is shown in Figure \ref{taxonomy}. Observing trajectory of development of swarm algorithms inspired in living being behaviors, there is a tendency where the number of new proposal is increasing (Figure \ref{algyear}).

An analysis was made considering references of this review to count algorithms proposed by authors and journals with more publications related with living beings behavior based inspired algorithms. Authors with more than two algorithms were considered inclusive mentioned as coauthors in original papers. Xin-She Yang (Author or involved in Cuckoo Search, Firefly Algorithm, Bat Algorithm, Flower Pollination Algorithm, Eagle Strategies, Virtual Ants and Bees, Catfish optimization algorithm) and Seyedali Mirjalili (Grey Wolf Algorithm, Whale Optimization Algorithm, Salp Swarm Algorithm, Ant Lion Optimization, Grasshopper Optimization Algorithm, Moth Flame Optimization Algorithm, Harris Hack Optimization) are the most prominent authors. Special mentions for Erik Cuevas (Side-Blotched Lizard Algorithm, Locust Swarm algorithm, Social Spider Algorithm, Yellow Saddle Goatfish), Jorge A. Ru\'iz -Vanoye (Rats Herd Algorithm, Feral Dogs Herd Algorithm, Dolphin Herd Algorithm, Zooplakton Swarm Algorithm), Kevin M. Passino (biomimicry of bacterial foraging, Honeybee social foraging algorithm), and Vijay Kumar (Seagull Optimization Algorithm, Spotted Hyena Optimization). Journal with more than four related publications are Applied Soft Computing, Advanced in Engineering Software,  Swarm and Evolutionary Computation, Soft Computing, Neural Computing and Applications,  International Journal of Bio-inspired Computation, The Scientific World Journal, Engineering Application of Artificial Intelligence, Expert System with Application and IEEE Transaction on Evolutionary Computation.

\section{Conclusion}

A new taxonomy is presented based on kingdom classifications and considering all known nature inspired living being algorithms. Several algorithms have been proposed in recent years, it makes a feeling that many of them do not  contribute to discipline or there is an abuse to promote new algorithms. Inclusive some author propose the use of only well established algorithms; by this way, they consider that it is not necessary to generate new algorithms only use better ones. Perhaps many algorithms could be useless, poorly tested or applied. Many of them have not tested intensively in different well known benchmarks; nevertheless, my particular opinion is that the doors should not close to new algorithms because they may contribute to new ideas, i.e. new behaviors for swarm robotics and perhaps appears a new revolutionary algorithm.

Finally, some authors are proposed an unifications of algorithms to establish universal algorithms, it is not clear because \emph{Not Free Lunch Theory} make not possible these condition but it has not be proven. In part because a mathematical foundation is required; however, in evolutionary or meta heuristics algorithms this problem persists until now. As a future work, highlighted algorithms based on behaviors of living forms could be included and make a study which includes memory used, computational complexity and parameters required. For this analysis a seudo-code of every algorithms must be generated in a homogenous way to allow comparisons between them. This is a monumental work,; however, we can reduce this work considering most highlighted algorithms of every kingdom.

\bibliographystyle{unsrtnat}
\bibliography{SwarmTaxonomy2020preprint}

\end{document}